\documentclass[sigconf, nonacm]{acmart}

\usepackage{multirow}
\usepackage[ruled]{algorithm2e}

\SetCommentSty{mycommfont}
\usepackage{soul}
\usepackage{array}
\newcolumntype{P}[1]{>{\centering\arraybackslash}p{#1}}
\usepackage{tikz}
\newcommand*\circled[1]{\tikz[baseline=(char.base)]{
            \node[shape=circle,draw,inner sep=1pt] (char) {#1};}}

\newcommand{\ours}[0]{DOM-LM}

\setcopyright{acmcopyright}
\copyrightyear{2018}
\acmYear{2018}
\acmDOI{10.1145/1122445.1122456}

\acmConference[Woodstock '18]{Woodstock '18: ACM Symposium on Neural
  Gaze Detection}{June 03--05, 2018}{Woodstock, NY}
\acmBooktitle{Woodstock '18: ACM Symposium on Neural Gaze Detection,
  June 03--05, 2018, Woodstock, NY}
\acmPrice{15.00}
\acmISBN{978-1-4503-XXXX-X/18/06}

\newcommand{\prashant}[1]{{\color{blue} [PS: #1]}}
\newcommand{\hs}[1]{{\color{cyan} [HS: #1]}}
\newcommand{\nop}[1]{}

\begin{document}

\title{DOM-LM: Learning Generalizable Representations\\ for HTML Documents}


\author{Xiang Deng}
\authornote{Work done when the first author was interning in Amazon.}
\email{deng.595@buckeyemail.osu.edu}
\affiliation{%
  \institution{The Ohio State University}
  \city{Columbus}
  \state{OH}
  \country{USA}
}

\author{Prashant Shiralkar}
\email{shiralp@amazon.com}
\affiliation{%
  \institution{Amazon.com}
  \city{Seattle}
  \state{WA}
  \country{USA}
}
\author{Colin Lockard}
\email{clockard@amazon.com}
\affiliation{%
  \institution{Amazon.com}
  \city{Seattle}
  \state{WA}
  \country{USA}
}
\author{Binxuan Huang}
\email{binxuan@amazon.com}
\affiliation{%
  \institution{Amazon.com}
  \city{Seattle}
  \state{WA}
  \country{USA}
}

\author{Huan Sun}
\email{sun.397@osu.edu}
\affiliation{%
  \institution{The Ohio State University}
  \city{Columbus}
  \state{OH}
  \country{USA}
}








\begin{abstract}
\nop{other candidate titles: \\
1. DOM-LM: Pre-training and Fine-tuning on DOM Tree Structured Documents. \\
2. DOM-LM: Learning Generalizable Representations for HTML documents\\
}
\nop{please see more comments under "review" mode:}
HTML documents are an important medium for disseminating information on the Web for human consumption. An HTML document presents information in multiple text formats including unstructured text, structured key-value pairs, and tables. Effective representation of these documents is essential for machine understanding to enable a wide range of applications, such as Question Answering, Web Search, and Personalization. Existing work has either represented these documents using visual features extracted by rendering them in a browser, which is typically computationally expensive, or has simply treated them as plain text documents, thereby failing to capture useful information presented in their HTML structure. We argue that the text and HTML structure together convey important semantics of the content and therefore warrant a special treatment for their representation learning. In this paper, we introduce a novel representation learning approach for webpages, dubbed DOM-LM, which addresses the limitations of existing approaches by encoding both text and DOM tree structure with a transformer-based encoder and learning generalizable representations for HTML documents via self-supervised pre-training. We evaluate DOM-LM on a variety of webpage understanding tasks, including Attribute Extraction, Open Information Extraction, and Question Answering. Our extensive experiments show that DOM-LM consistently outperforms all baselines designed for these tasks. In particular, DOM-LM demonstrates better generalization performance both in few-shot and zero-shot settings, making it attractive for making it suitable for real-world application settings with limited labeled data.

\end{abstract}




\maketitle

\begin{figure}
    \centering
    \includegraphics[width=0.95\linewidth]{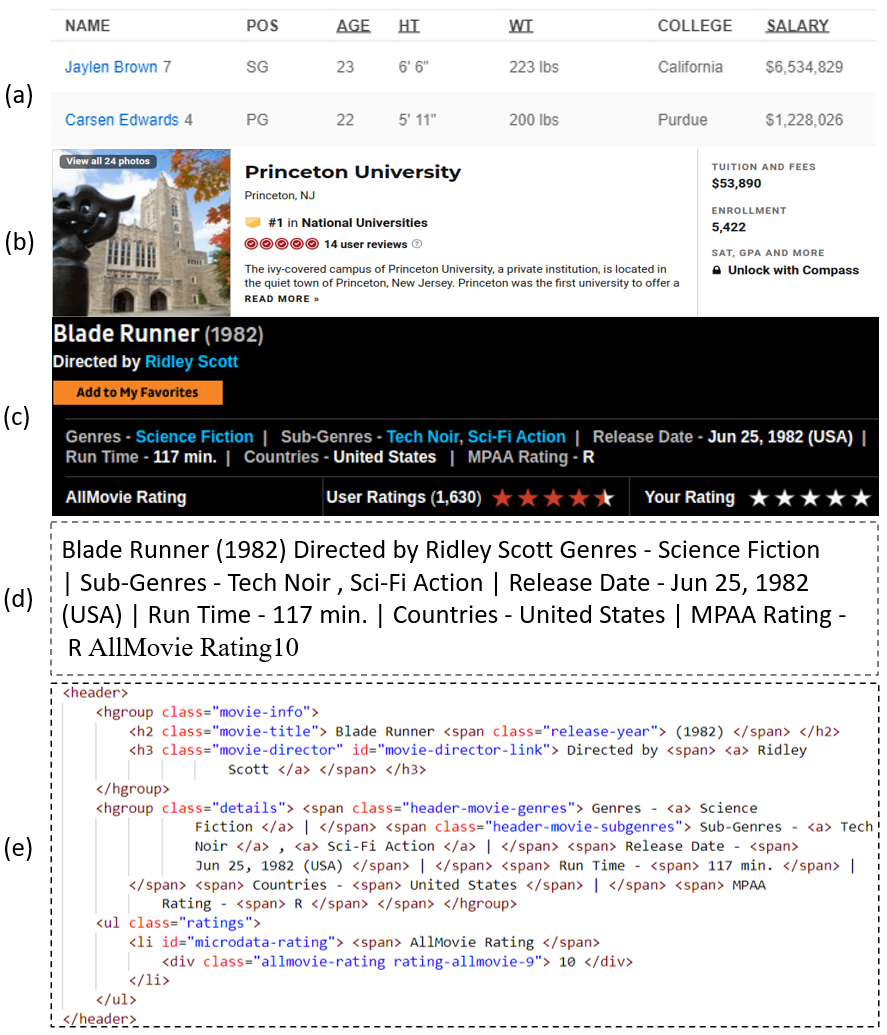}
    \vspace{-5pt}
    \caption{Illustration of different information formats on webpages (a, b, c). Figure (c), (d), (e) are the screen shot, textual content and source HTML document of the same page from IMDB, respectively. In (d), we can see that the layout structure is lost with only the textual content and it is merely a concatenation of short phrases without meaningful context. (e) shows all the essential DOM tree structure and markup information to properly render the page. Here, we use the HTML document in (e) as input. \nop{I like the diagram, but I think using diverse formats from the same (single) webpage might be clearer and can avoid confusion. The point we want to convey is not only that there are diverse formats, but that they together complement the textual context which can aid in learning better representations.}}
    \vspace{-15pt}
    \label{fig:intro}
\end{figure}
\section{Introduction}
\nop{This section needs more work especially on motivation, problem and challenges}
HTML documents are the primary source of information on the Internet. Owing to the wealth of such webpages, there have been long-standing efforts on tasks involving HTML documents, such as knowledge base construction~\cite{dong2014knowledge, wu2018fonduer}, question answering~\cite{kwiatkowski2019natural, chen2021websrc}, web search~\cite{Graupmann2004COMPASSAC, Escudeiro2009ExploringHT}, and recommendation~\cite{SnehaY2012APP}. HTML webpages are semi-structured. As shown in Figure~\ref{fig:intro}, they have information presented in multiple formats, including different levels of paragraphs, headers and titles, lists of key-value pairs and tables, and sorted according to the structure of the layout. Such complex structure makes it challenging to understand and utilize HTML documents, compared to plain text documents, which can be easily modeled as a sequence of tokens.

Take Figure~\ref{fig:intro} as an example. Figure~\ref{fig:intro} (c) is a partial screenshot of a webpage from IMDB. The webpage is rendered based on the source HTML document shown in Figure~\ref{fig:intro} (e), which can be parsed into a Document Object Model (DOM) tree. As we can see, the textual content in Figure~\ref{fig:intro} (d) is augmented with additional information from the DOM  tree structure and HTML markups. Using the text alone will lose the essential structural information, making it hard to infer the relationships between different HTML elements. \nop{Can you briefly give statistics from your analysis during internship that showed how much percent overlap and complementary information you found based on X pages of Y websites?}

Previous work~\cite{lockard-etal-2020-zeroshotceres, chen2021websrc} explores using visual features extracted from the rendered page, such as the bounding box coordinates of DOM nodes, or encoding of page screenshots. However, not only do these visual features incur additional computational cost to the model, but also the rendering process is time-consuming and requires extra space to store the necessary images, CSS, JavaScript files, making it hard to be applied in practice.\nop{Previous work has tried to directly utilize the underlying DOM tree, but primarily focuses on designing heavily-engineered heuristics or features. Early work mainly use the XPath for each node to extract extra structural features~\cite{gulhane2011web, freedom, lockard2018ceres}. Although such Xpath-based features perform well on the training site, they are usually brittle when generalizing to unseen websites that have different layout structures.}
More recently, SimpDOM~\cite{zhou2021simplified} directly utilizes the underlying DOM tree structure, while avoiding rendering-based features. It augments the node representation with relevant context retrieved from the DOM tree and has shown some promising results capturing website and domain invariant features. However, SimpDOM represents each node separately and relies on heuristics to construct the context. The model does not learn to perform tree-level contextualization, and the heuristics may fail to capture complex relationships among DOM nodes. How to effectively leverage the structural information inherent in the DOM tree remains a non-trivial problem. \nop{I think here we can summarize unique challenges faced in this setting that preclude us from using existing pre-trained models like BERT and its variants: 1) Can't apply BERT and variants as is since they only work on text. However, HTML documents are more than text; they contain structure and markup information that provide additional context to the textual semantics. 2) While BERT-like models can capture longer-range dependencies, they have limitations on input size and so cannot be applied to the linearized DOM tree in a straightforward manner without losing important information, thereby demanding a different strategy. (Edit: I see you have now added a paragraph below capturing this. Still would be useful to summarize first and then expand on the points in that paragraph.)}

Meanwhile, with the explosive growth and diversification of the web, there is an increasing need to develop a generalizable model for HTML document understanding.
Firstly, the model should be general so that it can benefit varied downstream tasks and save the effort on engineering task-specific features and architectures.
Secondly, the model needs to handle noisy multifarious webpages without requiring massive human annotations. This requires the model to not only adapt to the target website quickly, but also generalize well to websites that are not seen during training. How to learn such generalizable representations is another challenge.

\begin{figure*}
    \centering
    \includegraphics[width=0.95\linewidth]{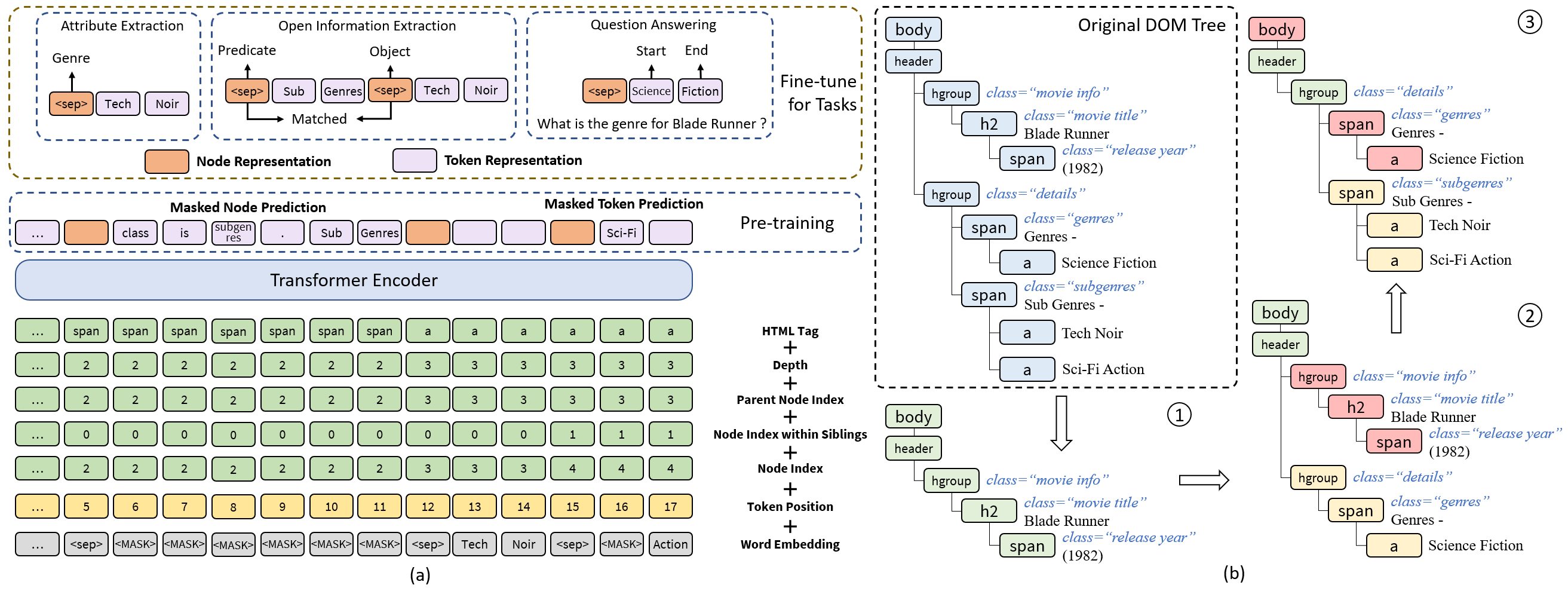}
    \vspace{-10pt}
    \caption{(a) shows the overall framework of \ours. We use extra position embeddings to model the DOM tree structure and encode the input with transformer encoder. The model is pre-trained with MLM objective by masking both individual tokens and entire nodes, and fine-tuned for downstream tasks with the learned node and token representations. We evaluate \ours\ on three tasks: attribute extraction, OpenIE and QA.
    (b) is an illustration of our DOM tree processor. Here we use a window of 5 and stride of 3\protect\footnotemark. At each step, yellow nodes are newly added, while red nodes will be pruned. The final subtree consists of the green and yellow nodes. }
    \vspace{-10pt}
    \label{fig:overview}
\end{figure*}

In this work, we present \ours, a novel representation learning approach for HTML documents. \ours\ tackles the aforementioned problems by encoding both text and DOM tree structure with a transformer-based encoder and pre-training the model with self-supervised objectives on unlabeled HTML documents.

Representation learning has achieved remarkable success in many fields such as natural language processing (NLP)~\cite{Peters2018DeepCW, devlin-etal-2019-bert, liu2019roberta} and computer vision~\cite{Chen2016InfoGANIR, Chen2020ASF}. Pre-trained with massive plain text documents, language models (LMs) such as BERT~\cite{devlin-etal-2019-bert} have become the foundation of most state-of-the-art NLP systems. BERT utilizes transformer~\cite{vaswani2017attention} to encode the input sentences, and pre-trains the model with masked language model (MLM) objective that predicts randomly masked tokens in the input. However, directly applying BERT to HTML documents may lead to unsatisfying results~\cite{zhou2021simplified}. The transformer encoder in BERT is designed for unstructured text and cannot handle structured input like HTML documents. Moreover, previous work has shown that MLM is weak at teaching the model to capture long-range dependencies~\cite{ram-etal-2021-shot,deng2021reasonbert}, not to mention the complex relationships between nodes in DOM tree.

In this work, we expand the transformer encoder in BERT with the ability to model structured HTML documents as DOM trees. Given a large DOM tree, we first split it into multiple subtrees that can fit into the encoder while retaining important tree-level context for each node. The subtrees are then linearized by concatenating the textual contents of each node as well as essential markup information from node attributes. We encode the linearized sequence with a structure-aware transformer encoder. Unlike a normal transformer encoder that uses one-dimensional incremental position ids to mark the order of tokens in a document, we use novel position embeddings such as index of the node, index of the parent node, depth in the tree and index within sibling nodes to incorporate the structural signals from the DOM tree into the input.

Following existing work~\cite{devlin-etal-2019-bert}, we pre-train the model with self-supervision to learn generalizable representations. However, conventional MLM objective only masks individual tokens and the model risks focusing on local cues. To encourage the model to do tree-level contextualization and to understand the layouts of webpages, we additionally mask entire DOM nodes, and force the model to recover all the masked tokens {in each masked node} by aggregating information from other nodes in the DOM tree. Meanwhile, we mask both node content and attributes to learn the use of HTML markups. We use the public available SWDE corpus~\cite{swde} for pre-training, which contains over 120k webpages from 80 representative websites (e.g. IMDB, ESPN) covering 8 domains.

We evaluate \ours\ on three downstream tasks: attribute extraction, open information extraction (OpenIE) and question answering (QA). On all three tasks, \ours\ consistently outperforms all baselines and achieves new state-of-the-art results. Moreover, \ours\ demonstrates superior generalization ability on few-shot and zero-shot experiments, where the model needs to adapt to the target website with a small number of training examples, or to generalize to websites that are not seen during training. For attribute extraction, \ours\ obtains an average of 94.2 F1 score on SWDE~\cite{swde} with around 10 webpages from each website for training. For OpenIE, \ours\ improves the Ceres~\cite{lockard-etal-2019-openceres, lockard-etal-2020-zeroshotceres} baselines by 24 and 4.7 F1 points under few-shot and zero-shot settings, respectively. Finally for QA, \ours\ outperforms the strong baseline~\cite{chen2021websrc} that leverages visual rendering features by 7.6 exact match points.\nop{We can perhaps give a one-line summary as a glimpse of our results (i.e \% improvement on each of the tasks w/ and w/o \ours{})}
\nop{\prashant{I think this paragraph is about details which can be omitted in the Introduction.} For Attribute extraction, we use the publicly available SWDE dataset \hs{You mentioned that the pre-training set is the same as this,  but without labels. Then our model would have seen the webpages used in testing set? Is it fair for other methods?} \prashant{I agree: needs justification or clarification}, and a new corpus we collected with a more diverse and noisier set of webpages to simulate real-world use cases. For OpenIE, we use the extended version of the SWDE data, which additionally annotates three verticals in SWDE with (subject, predicate, object) triples. For both attribute extraction and OpenIE, we focus on few-shot or zero-shot experiments with few or no labels available for the target site or vertical. This is closer to the real-world scenario where it is cost-prohibitive to have large amounts of training labels with the exponential growth of webpages.}

Overall, we make the following contributions through this work:
\begin{itemize}
    \item {We introduce a transformer-based representation learning approach for HTML documents called \ours\ that simultaneously encodes their textual and structural semantics, thereby extending the capabilities of existing language models for the modern Web.}
    \item We pre-train the model with self-supervision to learn generalizable representations, which have been confirmed to be effective in few-shot and zero-shot experiments.
    \item We experiment with three webpage understanding tasks: Attribute extraction, OpenIE and question answering. Our approach consistently outperforms the baselines on all tasks.
\end{itemize}

\section{Problem Definition}
We consider the problem of learning generalizable representations for HTML documents. We focus on webpages from semi-structured websites (e.g., IMDB) that present information in multiple text formats, including unstructured paragraphs, title and headers, lists of key-value pairs and tables. The structure and content of HTML documents can be represented with Document Object Model (DOM) as DOM trees. Each HTML element is associated with a DOM node, which encapsulates the textual content of the element and its HTML attribute markups. The pre-training task is to learn a representation for each DOM node and for each token of its textual content. 

We desire three characteristics in the learned representation: (1) It should capture textual semantics both local to a node and its surrounding tree-level context; (2) It should be task-agnostic so that it can benefit a wide range of downstream tasks, assuming sufficient overlap in their data distributions; (3) It should be robust to site-specific structural variations and thus can be generalized to a new website with minimal labeled data. \nop{Since we are learning representations for nodes and tokens, I think it would help to clarify what a DOM node is and what we mean by a token more concretely.}

\footnotetext{We count the number of nodes here for illustration. In real implementation, we count the total number of tokens in the nodes.}
\section{Proposed Approach}
We present \ours, a novel representation learning framework for HTML documents that employs a structure-aware encoder and self-supervised pre-training to learn textual semantics both local to a given node and that found in its DOM tree context. 
\nop{Here we should provide the key intuition and ideas behind our approach, a diagram depicting the architecture, and a high-level description of its its key components. We can provide details about the individual components in the sections below. You could start with something like: Our aim is to leverage the capacity of existing language model architectures like BERT for better pre-training for HTML documents... Then talk about what constitutes "context" in HTML setting, what position embedding means in this setting. Then introduce the key innovation: 1) tree transformation into constituent subtrees 2) novel position assignments for learning position embeddings.}

Inspired by the success of existing pre-trained language models for unstructured text documents such as BERT~\cite{devlin-etal-2019-bert}, our core idea is to extend their capabilities for HTML semi-structured documents by making them structure aware in addition to capturing their textual semantics.\nop{There are two main challenges to adapt BERT-like PLMs to structured HTML documents. (1) The vanilla transformer encoder in BERT is designed for plain texts, and cannot model the DOM tree structure in HTML documents. (2) PLMs such as BERT is only pre-trained on unstructured text documents, therefore unable to properly capture the tree-level context.} 
To accommodate the underlying DOM tree structure of HTML documents, we design \ours\ to consist of two parts: (1) A structure-aware encoder that processes the original DOM tree into a set of subtrees that each contain essential tree-level context, and models the tree structure with extra position embeddings, and (2) A self-supervised pre-training framework that pre-trains the model to capture textual semantics not only local to a DOM node, but also its tree-level context. We then fine-tune the pre-trained model for specific downstream tasks. We provide details on the pre-training and fine-tuning tasks in the following sections.

The overall design of \ours\ is shown in Figure~\ref{fig:overview} (a). The model architecture is general and can be applied to various downstream tasks. Most importantly, \ours\ learns generalizable representations via pre-training, which alleviates the need for large amounts of labeled data for each downstream task.

\subsection{Encoding the DOM Tree}
We consider the DOM tree representation of an HTML page. Here, each DOM node contains the textual contents and its associated HTML markup information, while the tree structure represents the layout indicating the relationship among different DOM nodes. How to encode such DOM trees as input to deep neural models is a key challenge for two reasons.

First, most existing text encoders are designed for unstructured textual sentences, and cannot handle the complex tree structure in DOM tree without special treatment. Second, in comparison to textual sentences, even a serialized version of a DOM tree can be considerably long to be used as input for most existing architectures, with number of DOM nodes often running in hundreds or even thousands. Thus, an arbitrary split is likely to result in a loss of important contextual information. Therefore, encoding DOM trees requires us to consider a more thoughtful strategy that retains the contextual information for each DOM node. 

We first look at how we carefully pre-process DOM trees to create plausible subtrees that can be individually encoded by our structure-aware encoder described in the following subsection.
\subsubsection{DOM tree Processor}

Our goal is to split a given DOM tree into a set of subtrees in a way that each subtree retains important context for tokens within it. We process the tree in two phases. 

First, we apply a cleaning phase that removes DOM nodes that have little influence on webpage structure and semantics such as \texttt{<script>} and \texttt{<style>}, keeping markups that serve as descriptions of a node such as \texttt{<class>} and \texttt{<id>}. After this cleaning phase, most DOM trees still may contain thousands of tokens, which can be hard to fit into existing deep neural models. 

As a second phase, we thus create subtrees by sliding a window with a fixed stride length, and thereby slicing the original tree into multiple overlapping pieces. However, simply sliding a window over the HTML document as done for text encoders may lose important tree-level context. For example, a child node may be far away from its parent node and may end up in separate subtrees. Instead, we design an algorithm that splits a DOM tree into multiple subtrees such that surrounding nodes important for understanding the semantics of a target node appear in the same tree.

We assume we have a fixed token budget $m$ to construct a subtree. We create our first subtree by traversing the DOM tree in pre-order and adding nodes\nop{with a stride length of $n$} until we reach the maximum number of allowed tokens $m$. We then create new overlapping subtrees by iteratively expanding and pruning the current subtree until we have examined all nodes in the original tree. The intuition is to create a new subtree from a current tree by inclusion of new nodes and their tokens that form the target nodes, while pruning any nodes that may be considered too far from the new target nodes and thus unlikely to constitute a useful context.

An example is shown in Figure~\ref{fig:overview} (b). Specifically, at each iteration, we first expand the subtree generated in previous iteration by adding new nodes with up to $n$ tokens in pre-order (yellow nodes in Figure~\ref{fig:overview} (b)). We then prune the expanded subtree by removing nodes (red nodes in Figure~\ref{fig:overview} (b)) in the following order to have a maximum subtree with $m$ tokens. First, we eliminate nodes that are far from the new nodes and are in different branches by a post-order traversal  (red nodes in the subtree \circled{2} of Figure~\ref{fig:overview} (b)). Second, if removing a node will split the tree, we remove the current root node of the subtree (\texttt{<body>} node in subtree \circled{3} of Figure~\ref{fig:overview} (b)). Finally, if this still splits the tree, we then remove the last node that was added. We provide full details of the subtree generation algorithm in Appendix~\ref{app:detail}. 

Given a subtree, we concatenate all nodes in pre-order, which is the order in which they appear in the HTML document. We then represent each DOM node as a concatenation of its HTML tag, attribute markups and textual content. \nop{I feel the reader can easily get lost in the tree-pruning details. However since this is key part of how we manage to create the right input, an example would help here. If space is a constraint, we can later even consider omitting the algorithm altogether, but an example would certainly help.}

\subsubsection{Structure-Aware Encoder}

As mentioned earlier, our idea is to extend the capabilities of existing language model architectures like BERT to capture both structural and textual semantics. To this end, we use the transformer encoder in BERT~\cite{devlin-etal-2019-bert} as our base encoder block. The backbone of the encoder is the principle of self-attention~\cite{vaswani2017attention}, where the model learns a contextualized representation for a token by aggregating information from other tokens in the input sequence (a subtree in our case). 

However, there are two key differences between unstructured text and our DOM setting, which require us to design novel position embeddings tailored for the DOM setting. First, unlike text setting, the tokens in our input tree are not all organized in a sequence. In addition to modeling the relationships between tokens sequence within a target node, we also need to aggregate information from tokens of other contextual nodes in the tree. Second, there are multiple ways in which the relative position of contextual tokens or nodes can be captured, thereby motivating us to consider distinct position embeddings for each distinct relative arrangement. 
\begin{table}[]
    \centering
    \caption{Various tree position features for a node/token.\nop{, each corresponding to one entry in the position matrix $P$.}}\vspace{-10pt}
    \resizebox{0.95\linewidth}{!}{
    \begin{tabular}{p{3cm}p{5cm}}
    \toprule
       Name  & Description  \\
       \midrule
        Node Index $P^0$ & Indicates tokens are in the same node.\\
        Parent Node Index $P^1$ & Indicates parent node, and helps differentiate sibling nodes.\\
        Node Index within Siblings $P^2$ & Indicates relative positioning of a sibling as preceding or succeeding.\\
        Depth $P^3$ & Depth in the DOM tree.\\
        HTML Tag $P^4$ & Type of node, e.g., \texttt{<h1>}, \texttt{<span>},  \texttt{<a>}.\\
        Token Position $P^5$ & Sequential position ids.\\
    \bottomrule
    \end{tabular}}\vspace{-5pt}
    \label{tab:tree_position}
\end{table}

We extend the one-dimensional position id $p_i$ in the text setting by considering a multi-dimensional position matrix $P_i=[P^0, P^1, ...,P^5]$ that indicates the position of token $i$ in the DOM tree, shown in Table~\ref{tab:tree_position}. In the following, we describe how we incorporate these novel positional embeddings in the existing self-attention framework.

In self-attention, attention score  $\boldsymbol{A}_{i j}$ between two tokens $i, j$ in the input is calculated as follows:
\begin{equation}
\boldsymbol{A}_{i j}= \boldsymbol{Q} \boldsymbol{K}^{T} = \mathbf{h}_{i} \boldsymbol{W}_{Q}\boldsymbol{W}_{K}^{T}\mathbf{h}_{j}^{T}.
\label{eqn:attention_score}
\end{equation}
Here $\mathbf{h}_i, \mathbf{h}_j$ are the hidden state output from the previous transformer layer or the input embedding layer, while $\boldsymbol{W}_{Q}$, $\boldsymbol{W}_{K}$ are parameter matrices. 

In a conventional transformer encoder, the initial $\mathbf{h}$ to the first layer is the sum of word embedding and position embedding, where $\mathbf{w}$ is a trainable word embedding for token $w \in \mathcal{W}$, and the position embedding is obtained by applying an encoding function $\phi: \mathbb{R} \rightarrow \mathbb{R}^{d}$ to each token's position $p_i$. 
\begin{equation}
    \mathbf{h}=\mathbf{w}+\phi(p)
\end{equation}
The attention score $\boldsymbol{A}_{i j}$ in Eqn.~\ref{eqn:attention_score} can be further factorized into:
\begin{equation}
{
\begin{aligned}
\boldsymbol{A}_{i j}=& (\mathbf{w}_{i}+\phi(p_i)) \boldsymbol{W}_{Q}\boldsymbol{W}_{K}^{T}(\mathbf{w}_{j}+\phi(p_j))^{T};\\
=& \underbrace{\mathbf{w}_{i} \boldsymbol{W}_{Q} \boldsymbol{W}_{K}^{T} \mathbf{w}_{j}^{T}}_{
\text { (a) } \boldsymbol{A}_{i j}^{\mathrm{ww}}
}+\underbrace{\mathbf{w}_{i} \boldsymbol{W}_{Q} \boldsymbol{W}_{K}^{T} \phi\left(p_{j}\right)^{T}}_{\text {(b) } \boldsymbol{A}_{i j}^{\mathrm{wp}}}+\underbrace{\phi\left(p_{i}\right) \boldsymbol{W}_{Q} \boldsymbol{W}_{K}^{T} \mathbf{w}_{j}^{T}}_{\text {(c) } \boldsymbol{A}_{i j}^{\mathrm{pw}}}\\
& +\underbrace{\phi\left(p_{i}\right) \boldsymbol{W}_{Q} \boldsymbol{W}_{K}^{T} \phi\left(p_{j}\right)^{T}}_{\text {(d) } \boldsymbol{A}_{i j}^{\mathrm{pp}}}.
\end{aligned}}
\end{equation}
As shown in \citet{zuegner_code_transformer_2021}, we can interpret the terms (a)-(d) as follows. (a) $\boldsymbol{A}_{i j}^{\mathrm{ww}}$ is the contribution from the semantic matching between the word embeddings of token $i$ and $j$. (b), (c), (d) take the token positions into consideration, and steer the attention towards certain positions based on the word embedding ($\boldsymbol{A}_{i j}^{\mathrm{wp}}$) or position ($\boldsymbol{A}_{i j}^{\mathrm{pp}}$) of token $i$, or bias towards word embedding based on the position of token $i$ ($\boldsymbol{A}_{i j}^{\mathrm{pw}}$). For unstructured texts, $p_i$ is just the position of a token in a sentence or document. In our context, we instead use position matrix described above.

\nop{I think the paragraphs above are all talking about the internals of BERT, right from Sec 4.2.2, which is not part of our contribution. So, the reader may be left wondering why she is shown all the details until they arrive at the following paragraph, which is brief. So, I suggest starting with a brief introduction of key parts of BERT and motivating what and why you plan to replace with novel components that are suited for the HTML setting. Provide intuition before introducing the concept.}


The new input embedding $\mathbf{h}_i$ passed to the first transformer layer is then calculated as
\begin{equation}
    \mathbf{h}=\mathbf{w}+\sum_{k=0}^{5}{\phi(P^k)}.
\end{equation}
In this work, we use lookup tables that store trainable embeddings as the encoding function $\phi$.

\subsection{Self-Supervised Pre-Training}

Since our structure-aware encoder design is based on transformer, we can reuse the parameters from existing pre-trained language models (PLMs) like BERT~\cite{devlin-etal-2019-bert} and RoBERTa~\cite{liu2019roberta}. However, these PLMs are pre-trained with only unstructured text documents. They may work well in modeling the textual semantics of individual DOM nodes, but cannot capture the structural and layout information from HTML DOM trees. To teach the model to understand HTML documents, we further pre-train the model on unlabeled webpages. 

We adopt the masked language model (MLM) objective from BERT, which randomly masks tokens in the input sequence and trains the model to recover them based on the context. One issue with MLM in our setting is that masking and predicting individual tokens may make the model focus only on local cues within a single DOM node. To encourage the model to learn to do tree-level contextualization and have a global view of the page layout, we additionally mask entire DOM nodes, so the model is forced to learn to recover the textual content of the masked node based on information aggregated from other nodes.

Specifically, we first select 15\% of the token positions at random, then iteratively sample DOM nodes and add all their tokens for masking until the masking budget (15\% of total tokens) has been spent. As in BERT, for a selected position, we (1) replace it with a special [MASK] token 80\% of the time, (2) replace it with another random token 10\% of the time, and (3) leave it unchanged otherwise.

Given a token position selected for MLM, which has a contexutalized representation $\mathbf{h}$ output by our encoder, the probability $P(w)$ of predicting its original token $w \in \mathcal{W}$ is then calculated as: 
\begin{align}
    P(w) &= \frac{\mathrm{exp}\left(h_w\cdot\mathbf{w}\right)}{\sum_{w_k \in \mathcal{W}}{\mathrm{exp}\left(h_w\cdot\mathbf{w}_k\right)}}\\
    \mbox{where } h_w&=\boldsymbol{W}_l\cdot \mathbf{h}+\boldsymbol{b}_l.
\end{align}\nop{LINEAR is not a formal term in a mathematical definition. Let's use something like $\boldsymbol{W}_l\cdot h+\boldsymbol{b}_l$ to replace it.}
Here $\boldsymbol{W}_{l}$, $\boldsymbol{b}_{k}$ are parameter matrices. Our final pre-training cross-entropy loss function is given as:
 \begin{equation}
     loss = \sum{\mathrm{log}\left(P(w)\right)},
 \end{equation}
 {where the summation is over all tokens selected individually or as part of the selected DOM nodes.}

\subsection{Fine-tuning for Downstream Tasks}

Our model architecture is general and can be applied to different downstream tasks by further fine-tuning its parameters and with minimal modifications. During fine-tuning, we use the output representation of \ours\ as input features to task-specific models and train them jointly with \ours. In this section, we demonstrate how we apply \ours\ to three representative webpage understanding tasks namely, attribute extraction, open information extraction and question answering. \nop{Say a bit about what fine-tuning means: we are going to update the pre-trained model parameters for each downstream task separately right? (IOW, weights aren't frozen, correct?)}

\subsubsection{Attribute Extraction}
Given a collection of homogeneously-templated webpages such as movie pages from IMDB.com, each describing a real-world entity (also sometimes called topic entity of the page), the attribute extraction task is to extract attribute values of a pre-defined set of attributes from a known ontology. 
%

We model the problem as a node classification task, in which we classify each DOM node into one of the attribute types or None if no attribute applies. 
We use the encoder output $\mathbf{h}$ associated with the HTML tags as the representations for DOM nodes. We then add a multilayer perceptron (MLP) classifier on top and compute a score $s=\text{MLP}(\mathbf{h})$ for each attribute type. The model is trained with cross-entropy loss. At inference time, we apply softmax to the output and predict the attribute with maximum score. 

While the attribute extraction task is well studied in information extraction literature, achieving high precision extractions under limited labeled data for an input website has been a scalability challenge and an open area of research. Thus, we are specifically interested in the use of \ours{} in zero to few-shot settings to push its boundaries further.

\subsubsection{Open Information Extraction}
Open information extraction (OpenIE) too aims at extracting facts about the topic entity from the HTML document. Unlike attribute extraction however, in OpenIE setting, there are no pre-defined attributes, and the task entails extracting all potential attribute (name, value) pairs from a webpage. 

We view this task as a binary node pair classification task. For a given pair of nodes $(i, j)$ from the DOM tree, we encode them with \ours{} to have their initial representations $\mathbf{h}_i, \mathbf{h}_j$. We then use a feed-forward network (FFN) with one layer to compute a score $s_p$ for node $i$ to represent an attribute name, $s_o$ for node $j$ to represent a value, $s_m$ for measuring compatibility of $i$ and $j$, and $s$ for $(i, j)$ being an (attribute, value) pair as follows:
\begin{gather}
s_p = \text{FFN}(\mathbf{h}_i);\ s_o = \text{FFN}(\mathbf{h}_j);\\
s_m =\ W_p\mathbf{h}_i\mathbf{h}_i^{T}W_o^T;\\
s = \text{FFN}([s_p;s_o;s_m]),
\end{gather}
where $W_p, W_o$ are trainable transform matrices. We train the attribute classifier, value classifier and pair classifier jointly with binary cross-entropy loss. At inference time, we apply a sigmoid function to $s_p, s_m, s$ and make extractions using a threshold of 0.5.

\subsubsection{Question Answering}
Question answering is the task of finding the answer for a given question from a HTML document. Instead of classifying entire nodes, the model needs to extract a textual span as the answer, which may be substring of a single node text, or scattered across multiple nodes. There are also yes/no questions, for which we prepend special \textit{yes} and \textit{no} tokens to the document. 

We concatenate the question and the HTML document as input to the encoder. Given the representation of each token in the HTML document $\mathbf{h}_i$ output by the encoder, we compute the score for token $i, j$ being the start and end token of the answer span as:
\begin{equation}
    s_s = \text{FFN}(\mathbf{h}_i);\ s_e = \text{FFN}(\mathbf{h}_j).
\end{equation}
The loss function for training is cross-entropy of the start and end token probabilities. At inference time, we extract the answer span by selecting the start and end locations with maximum scores.
\section{Experiments}
We conduct comprehensive experiments to answer the following research questions:

\begin{enumerate}
	\item Do the proposed structure-aware encoder and pre-training objectives help capture the structural information?
	\item Can \ours\ learn general representations for webpages that can benefit a wide range of tasks?
	\item Can \ours\ learn transferable representations that can generalize well across websites? \nop{For RQ3, it seems hard to link the claim of handling noisy real-world data to the experiment. Readers may not be aware that Demeter data is noisy and there is no experiments artificially add noise into the data.}
\end{enumerate}\nop{Briefly say how you plan to answer these RQs in the coming subsections.}
For the first question, we include two variants of \ours\ as baselines:\nop{should we briefly say why they make good baselines?} (1) \textbf{RoBERTa-Base}, which directly uses a pre-trained RoBERTa~\cite{liu2019roberta} encoder. (2) \textbf{RoBERTa-Structural}, which uses the same structure-aware encoder as \ours, but directly adopts the weights from pre-trained RoBERTa without further pre-training. 

For the second question, we experiment with three different tasks: attribute extraction, open information extraction (OpenIE) and question answering. For attribute extraction, the task is to extract potential values for a given attribute from the webpage. The model needs to capture both semantics from local textual content and tree-level context. For OpenIE, the task is to extract all potential attribute (name, value) pairs from the webpage without a pre-defined ontology. Thus the model needs to understand the layout structure and markup information to identify all potential mentions of predicates and object values, and figure out the relationships between them to match them together. Finally, question answering is the task of extracting the answer span from the webpage for a given question, and requires the model to jointly understand the question and the webpage content. 

For the third research question, we focus on few-shot and zero-shot experiments. In the few-shot setting, the model is trained and evaluated on the same set of websites, but we only use a small sample of training instances. In the zero-shot setting, the model is trained and evaluated on different sets of websites. These settings evaluate the model's ability to learn generic site-agnostic knowledge, which can be transferred to new websites in the same or different domain. Such ability is crucial for scalability to circumvent the challenge of labeled data for every new website.

\subsection{Datasets}
We use the following datasets summarized in Table~\ref{tab:data}.
\vspace{0.5em}
\begin{table}[]
    \centering
    \caption{Data statistics. \# Nodes is the average number of DOM nodes in the HTML documents after cleaning noisy HTML tags. For Movie-Music, we also show the number of clusters in parenthesis, which is the actual number of unique webpage templates.}\vspace{-10pt}
    \resizebox{\linewidth}{!}{
    \begin{tabular}{lP{2.5cm}P{2cm}P{2cm}}
    \toprule
    Dataset&SWDE~\cite{swde, lockard-etal-2019-openceres}&Movie-Music&WebSRC~\cite{chen2021websrc}\\
    \midrule
    Usage& Pre-train, Attribute extraction, OpenIE&Attribute extraction&Question Answering\\
    \# Pages &124,291&2,954&5,462\\
    \# Domains &8&2&10\\
    \# Websites &80&96 (612)& 60\\
    \# Nodes&563.6&706.5&155.6\\
    \bottomrule
    \end{tabular}}\vspace{-15pt}
    \label{tab:data}
\end{table}

\begin{table*}[t]
    \centering
    \caption{Few-shot attribute extraction results on SWDE (F1). We use 0.5\% of data for training. \nop{what is the reported metric in these tables? Please add it in the caption.}}\vspace{-10pt}
    \resizebox{0.9\textwidth}{!}{
    \begin{tabular}{lP{1.3cm}P{1.35cm}P{1.35cm}P{1.35cm}P{1.35cm}P{1.35cm}P{1.35cm}P{1.35cm}P{1cm}}
    \toprule
    Model&Movie&Auto&Camera&Restaurant&Book&Job&NBAplayer&University&\bf AVG  \\
    \midrule
    SimpDOM~\cite{zhou2021simplified}& 69.8& 84.8& 54.2& 85.4& 71.0& 80.2& 65.4& 89.6&\bf 75.1\\
    RoBERTa-Base     & 92.1& 96.8& 83.5& 97.2& 93.0& 92.8& 84.4& 97.9&\bf 92.2\\
    RoBERTa-Structural     & 93.0& 97.1& 83.9& 97.3& 93.2& 93.1& 86.4& 98.0&\bf 92.7\\
    \ours     & 94.6& 97.7& 87.7& 97.7& 94.8& 94.8& 87.9& 98.3& \bf 94.2\\
    \bottomrule
    \end{tabular}}\vspace{-7pt}
    \label{tab:few_closedie}
\end{table*}

\begin{table*}[t]
    \centering
    \caption{Zero-shot attribute extraction results on SWDE (F1). We use 10\% of data from 2 or 5 seed websites for training.}\vspace{-10pt}
    \resizebox{0.9\textwidth}{!}{
    \begin{tabular}{clP{1.3cm}P{1.35cm}P{1.35cm}P{1.35cm}P{1.35cm}P{1.35cm}P{1.35cm}P{1.35cm}P{1cm}}
    \toprule
    \# Seed&Model&Movie&Auto&Camera&Restaurant&Book&Job&NBAplayer&University&\bf AVG  \\
    \midrule
    \multirow{4}{*}{2}&SimpDOM~\cite{zhou2021simplified}&34.6&47.2&44.6&43.4&34.6&43.4&28.8&59.8&\bf 42.1\\
    &RoBERTa-Base     & 49.3&42.7&48.8&54.7&48.3&46.7&34.2&36.6&\bf 45.2\\
    &RoBERTa-Structural     & 47.7&45.4&52.8&56.6&51.9&52.1&48.6&46.5&\bf 50.2\\
    &\ours     & 71.9&64.9&73.3&79.0&64.8&72.4&64.8&68.0&\bf 69.9\\
    \midrule
    \multirow{4}{*}{5}&SimpDOM~\cite{zhou2021simplified}&
    47.6&56.0&47.6&58.6&38.0&51.6&58.6&70.0&\bf 53.5\\
    &RoBERTa-Base     & 69.1&63.3&65.5&73.9&61.1&59.3&60.4&67.2&\bf 65.0\\
    &RoBERTa-Structural     & 71.2&60.6&65.5&81.5&63.6&67.0&69.8&73.1&\bf 69.0\\
    &\ours     & 80.7&74.9&75.3&83.4&70.8&78.4&75.1&81.2&\bf 77.5\\
    \bottomrule
    \end{tabular}}\vspace{-7pt}
    \label{tab:zero_closedie}
\end{table*}

\begin{table}[]
    \centering
    \caption{Few-shot and zero-shot attribute extraction results on Movie-Music (F1). We use 60\% of pages (few-shot, around 3 pages per cluster) or clusters (zero-shot) for training. \nop{Since Demeter is an Amazon internal team name, maybe we should change it to other names like Movie and Music.}}\vspace{-10pt}
    \resizebox{0.82\linewidth}{!}{
    \begin{tabular}{lcccc}
    \toprule
        \multirow{2}{*}{Model} & \multicolumn{2}{c}{Few-shot} & \multicolumn{2}{c}{Zero-shot}\\
        & Movie & Music &Movie&Music\\
        \midrule
        SimpDOM~\cite{zhou2021simplified}&36.0&25.2&28.6&19.8\\
        RoBERTa-Base&73.8&66.8&44.2&48.4\\
        RoBERTa-Structural&74.2&69.4&46.7&47.6\\
        \ours&76.1&71.5&47.5&52.6\\
        \bottomrule
    \end{tabular}}\vspace{-12pt}
    \label{tab:demeter}
\end{table}

\noindent\textbf{SWDE.} SWDE~\cite{swde}, originally assembled to target attribute extraction task, includes over 120k webpages covering 8 domains with 3 to 5 attributes of interest per domain. Each domain has 10 representative websites, such as IMDB for Movie, ESPN for NBAplayer and usnews for University. \citet{lockard-etal-2019-openceres} extend SWDE for OpenIE, additionally annotating all (subject, predicate, object) triples for 21 websites in 3 domains. SWDE originally contains 4,480 object annotations for 3 predicates from these 21 websites on average, while \citet{lockard-etal-2019-openceres} extract an average of 41k triples for 36 predicates. The released data only contains annotated values. Following \citet{lockard-etal-2020-zeroshotceres}, we link values to specific DOM nodes using established string matching techniques. However, the NBAplayer domain has many numerical relations and pages with large tables containing numbers, causing considerable ambiguity during evaluation. For clear evaluation, we perform OpenIE experiments only on the Movie and University domains.

\vspace{0.1em}
\noindent\textbf{Movie-Music.} To further test how well \ours\ can handle multifarious noisy webpages in real-world, we collect a new dataset from 96 websites in Movie and Music domains. We first segregate webpages adhering to distinct HTML templates by applying a url-based clustering algorithm~\cite{cluster}. As we are interested in the few-shot and zero-shot results, we sample 5 webpages from each cluster and annotate their attributes. In the end, we have 2,954 webpages from 612 clusters with 250 attributes. Although our new dataset is much smaller than SWDE, it is more diverse in terms of unique layouts and attribute types. Note SWDE contains 80 template clusters corresponding to the 10 websites in 8 domains.

\vspace{0.1em}
\noindent\textbf{WebSRC.} WebSRC~\cite{chen2021websrc} consists of 440K question-answer pairs collected from 6.5K webpages covering 10 domains. The authors only released the training and development dataset, which contains 5,462 webpages from 60 websites. The HTML documents used in WebSRC are snippets of the original webpage, so the average number of DOM nodes is much smaller compared to other two datasets. 

\subsection{Implementation Details}
We pre-train \ours\ on SWDE since it is the largest among all the datasets we used.  For pre-training, we use all the raw HTML documents without any human labels. We implement the encoder based on RoBERTa-Base using Huggingface Transformers~\cite{wolf-etal-2020-transformers}, which has 12 layers and hidden size of 768. The model is initialized with pre-trained RoBERTa-Base, and further pre-trained on SWDE for 5 epochs with batch size of 24, using a linear learning rate scheduler with max learning rate of 1e-4 and warm up for the first half epoch.

We use a 2 layer MLP with hidden size of 768 for classification in attribute extraction. For OpenIE and QA, we use single layer feedforward networks. During fine-tuning, parameters of these prediction heads are updated together with the pre-trained encoder. We use batch size of 16, learning rate of 3e-5 to fine-tune our models for all downstream tasks. The total number of training epochs varies for different datasets, please check Appendix~\ref{app:detail} for more details.
\nop{I think it would help to make clear whether data for fine-tuning was part of pre-training as well. This was the confusing part during first reading. For a new reader, it may not be clear.}
\nop{Another baseline I think maybe worth adding here is RoBERTa-Base-pretrain, which pretrains a vanilla RoBERT encoder on the HTML corpus. Otherwise people may argue that we can treat HTML file as normal text and it may only need some pretraining adaptation.}
\nop{One thing missing here is the model hyperparameter settings.}

\subsection{Attribute Extraction}
\noindent\textbf{Experimental Setup.} In the few-shot setting for SWDE, we take 0.5\% of data (around 10 pages from each website) for training and divide the rest into development and testing with a 50:50 ratio. In the zero-shot setting, we take 2/5 seed websites from each domain, and sample 10\% of pages from each seed website for training. For the remaining websites, one is used for development and the rest are used for testing. For the new Movie-Music corpus we collected, we use 3 pages from each cluster for training, 1 for development and 1 for testing under the few-shot setting. For the zero-shot setting, we split the clusters into training, development and testing follow a 60:20:20 ratio. We also include SimpDOM~\cite{zhou2021simplified} as a baseline, which is a state-of-the-art extraction model. We use their code released on GitHub\footnote{\url{https://github.com/google-research/google-research/tree/master/simpdom}} and experiment with our settings. Also note that for all models, we do not apply extra post-processing such as site-level voting in original SimpDOM~\cite{zhou2021simplified}.

\vspace{0.1em}
\noindent\textbf{Results.} We evaluate the model with attribute value-level F1 scores. Unlike the page-level F1 scores used in previous work~\cite{zhou2021simplified, freedom} that scores the extraction on a page as success as long as one of the predictions is correct, attribute value-level F1 considers the case where there are multiple values for the same attribute in a webpage, and penalizes the false positive predictions made by the model\footnote{For page-level F1 scores please check Appendix~\ref{app:more_results}}. Results are summarized in Table~\ref{tab:few_closedie}, \ref{tab:zero_closedie} and \ref{tab:demeter}. 

First we can see that our RoBERTa-Base variant already consistently outperfoms SimpDOM. The heuristic used by SimpDOM to retrieve relevant context for each node from the DOM tree seems to fail to generalize well on the real-world Movie-Music dataset we collect and results in much lower F1 scores on it than SWDE. On the other hand, \ours\ including the two variants still have reasonable performance, which demonstrates that our DOM tree processor is able to properly split the tree and provide essential context in the input for the encoder. RoBERTa-Structural and \ours\ gradually improve the performance over RoBERTa-Base and obtain the new state-of-the-art results, showing the advantage of applying our structure-aware encoder and pre-training on structured HTML documents. In the few-shot setting, we can see that \ours\ obtains 94.2 F1 score on SWDE with only around 10 pages in each website for training. The advantage of pre-training is more clear in zero-shot experiments, where \ours\ outperforms the second-best model by a large margin. For the Movie-Music dataset, most of the websites are not included in pre-training, and the entire Music domain is not covered, while \ours\ still improves the performance. This shows that our pre-trained representations are transferable, and can generalize well across websites and domains.

\subsection{OpenIE}
\begin{table}[]
    \centering
    \caption{Few-shot and zero-shot OpenIE results (Pair F1). We use 0.5\% of data for training under the few-shot setting. For the zero-shot setting, the model is trained 10\% of data, with 6 and 3 seed websites for Movie and University, respectively.}\vspace{-10pt}
    \resizebox{0.9\linewidth}{!}{
    \begin{tabular}{lcccc}
    \toprule
        \multirow{2}{*}{Model} & \multicolumn{2}{c}{Few-shot} & \multicolumn{2}{c}{Zero-shot}\\
        & Movie & University &Movie&University\\
        \midrule
        OpenCeres~\cite{lockard-etal-2019-openceres}&77&40&-&-\\
        ZeroShotCeres~\cite{lockard-etal-2020-zeroshotceres}&-&-&50&50\\
        RoBERTa-Base&83.0&68.6&35.6&38.0\\
        RoBERTa-Structural&85.4&73.0&39.9&42.3\\
        \ours&87.5&77.5&54.1&55.2\\
        \bottomrule
    \end{tabular}}\vspace{-12pt}
    \label{tab:openie}
\end{table}
\noindent\textbf{Experimental Setup.} In the few-shot setting, we again show the results with 0.5\% of webpages for training. In the zero-shot setting, for each domain, we use all but two websites for training, and one for development, one for testing. We compare with the Ceres baselines~\cite{lockard-etal-2019-openceres, lockard-etal-2020-zeroshotceres} for OpenIE, and use the number reported in \citet{lockard-etal-2020-zeroshotceres}. However, their settings are different from ours. In the few-shot setting, we compare with OpenCeres~\cite{lockard-etal-2019-openceres}, where they use distant supervision to obtain training labels. Their training instances contain more noise but also are much larger than the 10 pages use. In the zero-shot setting, we compare with ZeroShotCeres~\cite{lockard-etal-2020-zeroshotceres}. However, their results are obtained by training the model with all sites but one, spanning Movie, NBA, and University, and test on the held-out set.

\vspace{0.1em}
\noindent\textbf{Results.}
Results for OpenIE are shown in Table~\ref{tab:openie}. we follow the “lenient” scoring method introduced by \citet{lockard-etal-2019-openceres}, which considers an extraction as correct if the relation string matches any of acceptable surface forms listed by the ground truth for that object. We can see that on the same set of webpages from SWDE, OpenIE is more challenging than attribute extraction. Using the structure-aware encoder alone (RoBERTa-Structural) does not outperform ZeroShotCeres, this is probably because ZeroShotCeres uses Graph Neural Network that is also capable of modeling the tree structure and leverages extra visual features from rendering. Nevertheless, \ours\ outperforms the Ceres baselines under both settings, even though our models are trained with fewer training instances, demonstrating the usefulness of self-supervised pre-training.

\begin{table}[]
    \centering
    \caption{QA results. Since the test set for WebSRC~\cite{chen2021websrc} has not been released, here we report results on the development set.}\vspace{-10pt}
    \resizebox{0.75\linewidth}{!}{
    \begin{tabular}{lcccc}
    \toprule
     \multirow{2}{*}{Model}    &\multicolumn{2}{c}{1\%}&\multicolumn{2}{c}{Full}  \\
     & EM&F1&EM&F1\\
     \midrule
    T-PLM (BERT-Base)~\cite{chen2021websrc}&-&-&52.1&61.6\\
    H-PLM (BERT-Base)~\cite{chen2021websrc}&-&-&61.5&67.0\\
    V-PLM (BERT-Base)~\cite{chen2021websrc}&-&-&62.1&66.7\\
    RoBERTa-Base     &55.8&61.7&65.7&70.1 \\
    RoBERTa-Structural&58.7&64.0&68.9&73.2\\
    DOM-LM&61.3&67.3&69.7&73.9\\
    \bottomrule
    \end{tabular}}\vspace{-10pt}
    \label{tab:qa}
\end{table}
\subsection{Question Answering}
\noindent\textbf{Experimental Setup.}
We use the WebSRC~\cite{chen2021websrc} dataset for question answering. We use the official split that splits the datasets to training and development sets at the website level. This is the same as our zero-shot setting that evaluates the model on unseen websites. In addition to using all 307,315 QA pairs for training as \citet{chen2021websrc} (Full in Table~\ref{tab:qa}), we experiment with only 1\% of QA pairs for training. We compare with three baselines proposed by \citet{chen2021websrc} that have similar size as ours: T-PLM that only uses textual content of the HTML document, H-PLM that keeps the HTML tags, and V-PLM that incorporates visual features.

\vspace{0.1em}
\noindent\textbf{Results.} Results are summarized in Table~\ref{tab:qa}. Following common QA practice, we evaluate with exact match (EM) and answer F1 score. We can see that our RoBERTa-Base variant already outperforms all baselines, possibly because our DOM tree processor is better at preserving the DOM tree context in the input. \ours\ and RoBERTa-Structural achieve similar performance under the full data setting where there are abundant examples to fine-tune the model. But it is still clear in the few-shot setting that pre-training helps to learn representations that can quickly adapt to the downstream task. Meanwhile, QA requires joint understanding of the question and HTML document. The superior performance of \ours\ shows that our pre-training does not hurt the text understanding capability of the model that inherits from pre-trained RoBERTa-base.

\section{Related Work}
\noindent\textbf{Language Model Pre-training.}
Representation learning for plain text documents is a long-studied problem. The most prevalent approaches are variants of pre-trained language models such as BERT~\cite{devlin-etal-2019-bert} and RoBERTa~\cite{liu2019roberta}. These PLMs pre-train text encoders like transformer~\cite{vaswani2017attention} with self-supervised objectives and have shown promising results on natural language processing tasks. BERT first introduced the masked language model pre-training objective, which required the model to recover masked tokens in the input so as to learn syntactic and semantic characteristics of word use. Some recent work also explores representation learning for knowledge base~\cite{Lin2015ModelingRP, Hao2019UniversalRL} and web tables~\cite{Deng2020TURLTU, Herzig2020TaPasWS}, which sheds some light on extending language model pre-training to structured data. Concurrent to our work, MarkupLM~\cite{li2021markuplm} also targets pre-training for HTML documents. Our work differs from theirs in two respects. First, they take the raw HTML document as input and utilize the DOM tree with XPath embeddings, whereas we focus on processing the tree structure first to only retain essential tree-level context in the input, and then using multiple tree position embeddings to incorporate the structural signals. Second, we aim at learning a generalizable representation that can be effective especially in zero-shot and few-shot settings.

\noindent\textbf{Information Extraction from HTML Documents.}
There are two main regimes for information extraction from HTML documents: traditional attribute extraction, where the model only looks for values of attributes in a pre-defined ontology, and open information extraction, where the model needs to identify values for all attributes, even ones that do not exist in the ontology. Early approaches on information extraction involved learning rule-based extractors~\cite{gulhane2011web, lockard2018ceres}, heavily relying on hand-crafted rules or labels, and thus making them brittle in practice and not scalable when the goal is to extract from a large number of HTML templates. FreeDOM~\cite{freedom} is a recent deep learning approach that learns a representation for each DOM node using both its text and markup information, capturing long-range dependencies between nodes using a relational neural network. SimpDOM~\cite{zhou2021simplified} leverages the structural information by enriching the node representation with useful contexts retrieved from the DOM tree. Recently, there has been some work~\cite{lockard-etal-2020-zeroshotceres} utilizing visual features extracted from rendered webpages to learn more generalizable features. However, these methods either use heavily engineered or computationally expensive features that can be hard to scale in real-world applications, or rely on manual heuristics to construct the context for each DOM node, which may fail to capture complex relationships among DOM nodes. In contrast, we directly encode the DOM tree with a structure-aware encoder and use self-supervised pre-training to learn generalizable representations. \nop{We should compare/contrast these briefly with our approach. O/w the para is without a point, just a summary of related work.}

\noindent\textbf{Question Answering.}
Question answering (QA) has become an important testbed for machine understanding evaluation. Recent advances in PLMs have greatly improved the performance of QA models on benchmark datasets with only plain text passages. Recently, \citet{chen2021websrc} introduced the WebSRC dataset, where the answer resides in a given webpage. The dataset poses new challenges for the model to capture the structural information in the webpage, as the text extracted from the raw HTML documents are just short phrases without context. \citet{chen2021websrc} directly used PLMs pre-trained for unstructured text to encode the tags and tokens from the HTML document, and relied on visual representations learned from the rendered webpage to capture the structural information. Although their method outperforms text-based QA model by a large margin, creating such visual features is computationally expensive and rendering the webpage is not always feasible in practice as images, CSS, JavaScript files are prone to be out-of-date.

\section{Conclusion}
In this work, we introduce a transformer-based representation learning approach for HTML web documents called \ours \ that simultaneously encodes their textual and structural semantics, thereby extending the capabilities of existing language models for the modern Web. The model is pre-trained with self-supervision to learn generalizable representations. Extensive experiments in three downstream tasks show that \ours \  consistently outperforms all baselines designed for these tasks. In particular, \ours \ demonstrates better generalization performance both in few-shot and zero-shot settings, making it attractive for use toward real-world applications aiming to leverage vast amounts of information on the Web.


\bibliographystyle{ACM-Reference-Format}
\bibliography{ref.bib}

\clearpage
\appendix
\section{Implementation Details}
\label{app:detail}
The detailed algorithm for our DOM tree processor is shown in Algorithm~\ref{alg:dom}. In Table~\ref{tab:hyper} we provide the total number epochs used in fine-tuning for different experiments.
\section{More results}
\label{app:more_results}
We show the page-level F1 scores for attribution extraction in Table~\ref{tab:few_closedie_page} and \ref{tab:zero_closedie_page}. Here we calculate the scores follow the script released by \citet{zhou2021simplified}. Page-level F1 considers the extraction for a webpage a success as long as one of the predictions made by the model matches with the ground truth. This ignores the false positives made by model, thus much higher than fact-level F1 scores.

\begin{algorithm}[t]
\nop{Put comments for major snippets; otherwise can be hard to parse.}
\caption{DOM tree Processor to generate subtrees.}\label{alg:dom}
\SetKw{Break}{break}
\KwIn{DOM tree $T$, Max number of tokens $M$, Stride $S$;}
\KwOut{List $L$ where each item is a subtree $t \in T$;}
Initialize $L$ as an empty list\;
Preorder and Postorder traverse $T$, store the visiting order in $N_{pre}, N_{post}$\;
Initialize $New$ as an empty list to store new nodes visited in each turn\;
\tcc{Initialize the first subtree.}
\For{each node $x\in N_{pre}$}{
    \eIf{Total number of tokens for nodes in $New > M$}{\Break}{Add $x$ to $New$\;}
}
\end{algorithm}

\begin{table}[]
    \centering
    \caption{Total number of training epochs for different experiments. We use linear learning rate scheduler and warmup for the first one tenth part of training (Warm up for 2 epochs if the total number of epochs is 20).}
    \begin{tabular}{lc}
    \toprule
    Experiment&Epoch\\
    \midrule
    Attribute Extraction - SWDE \\
    Few-shot&20\\
    Zero-shot (2 seed)&20\\
    Zero-shot (5 seed)&10\\
    \midrule
    Attribute Extraction - Movie-Music \\
    Few-shot&5\\
    Zero-shot&5\\
    \midrule
    OpenIE\\
    Few-shot&50\\
    Zero-shot (Movie)&20\\
    Zero-shot (Music)&50\\
    \midrule
    QA\\
    Full&2\\
    1\% data&10\\
    \bottomrule
    \end{tabular}
    \label{tab:hyper}
\end{table}

\makeatletter
\newcommand{\removelatexerror}{\let\@latex@error\@gobble}
\makeatother
{\begingroup
\removelatexerror
\begin{algorithm}[H]
\SetKw{Break}{break}
\tcc{Continue Algorithm 1.}
\While{$New$ is not empty}{
$Visited \leftarrow$ List of nodes in $N_{pre}$ that ordered before nodes in $New$\;
$Length \leftarrow$ Total number of tokens for nodes in $New$ and $Visited$\;
\tcc{Prune the subtree in postorder.}
\For{each node $x\in N_{post}$}{
    \eIf{$x$ in $New$, or $Length < M$}{\Break}{
    Remove $x$ from $Visited$\;
    $Length\ -=$ Number of tokens in $x$\;
    }
}
\tcc{Prune the root node if this will not split the tree, otherwise prune the last node that got added.}
\While{$Length > M$}{
$x_{root} \leftarrow Visited[0]$\;
$NumChild \leftarrow$ Number of children of $x_{root}$ that is in $New$ or $Visited$\;
\eIf{$NumChild < 2$}{
$Length\ -=$ Number of tokens in $x_{root}$\;
remove $x_{root}$ from $Visited$\;
}{
$x_{last} \leftarrow New[-1]$\;
$Length\ -=$ Number of tokens in $x_{last}$\;
remove $x_{last}$ from $New$\;
}
}
Create subtree $t$ with nodes in $New$ and $Visited$ and add to $L$\;
$x_{last} \leftarrow New[-1]$\;
\tcc{Slide the window and expand the subtree.}
Empty $New$\;
\For{each node $x \in N_{pre}$ that is after $x_{last}$}{
\eIf{Total number of tokens for nodes in $New > S$}{\Break}{Add $x$ to $New$\;}
}
}
\end{algorithm}
\endgroup}

\begin{table*}[t]
    \centering
    \caption{Few-shot attribute extraction results on SWDE (page-level F1). The model is trained with 0.5\% of data for each domain. \nop{what is the reported metric in these tables? Please add it in the caption.}}
    \resizebox{0.95\textwidth}{!}{
    \begin{tabular}{cP{1.35cm}P{1.35cm}P{1.35cm}P{1.35cm}P{1.35cm}P{1.35cm}P{1.35cm}P{1.35cm}P{1cm}}
    \toprule
    Model&Movie&Auto&Camera&Restaurant&Book&Job&NBAplayer&University&\bf AVG  \\
    \midrule
    SimpDOM~\cite{zhou2021simplified}& 87.4&98.8&81.8&94.8&88.0&95.2&89.4&98.0&\bf 91.7\\
    RoBERTa-Base     & 99.9&99.9&97.7&99.8&99.7&99.7&98.4&99.9&\bf 99.4\\
    RoBERTa-Structural     & 99.9&100.0&98.0&99.9&99.8&99.8&99.6&99.7&\bf 99.6\\
    \ours     & 100.0&100.0&99.4&100.0&99.9&99.9&99.8&100.0&\bf 99.9\\
    \bottomrule
    \end{tabular}}
    \label{tab:few_closedie_page}
\end{table*}

\begin{table*}[t]
    \centering
    \caption{Zero-shot attribute extraction results on SWDE (page-level F1). The model is trained with 10\% of data from 2 or 5 seed websites for each domain.}
    \resizebox{0.95\textwidth}{!}{
    \begin{tabular}{ccP{1.35cm}P{1.35cm}P{1.35cm}P{1.35cm}P{1.35cm}P{1.35cm}P{1.35cm}P{1.35cm}P{1cm}}
    \toprule
    \# Seed&Model&Movie&Auto&Camera&Restaurant&Book&Job&NBAplayer&University&\bf AVG  \\
    \midrule
    \multirow{4}{*}{2}&SimpDOM~\cite{zhou2021simplified}&71.0&65.6&76.2&74.8&61.0&59.0&68.2&82.8&\bf 69.8\\
    &RoBERTa-Base     & 71.6&65.9&77.1&73.3&68.8&60.3&61.1&56.0&\bf 66.8\\
    &RoBERTa-Structural     & 73.1&68.2&81.8&77.4&77.3&63.4&74.7&63.3&\bf 72.4\\
    &\ours     & 91.0& 87.4& 95.1& 98.0& 90.5& 84.8& 91.1& 95.5&\bf 91.7\\
    \midrule
    \multirow{4}{*}{5}&SimpDOM~\cite{zhou2021simplified}&
    83.0&77.2&88.6&80.2&72.8&71.0&89.6&93.2&\bf 82.0\\
    &RoBERTa-Base     & 87.3&81.6&87.7&88.0&85.6&78.1&85.7&83.9&\bf 84.7\\
    &RoBERTa-Structural     & 89.9&84.2&92.0&92.2&87.2&83.0&94.5&89.0&\bf 89.0\\
    &\ours     & 94.4&92.2&97.3&98.0&95.8&93.4&95.3&99.2&\bf 95.7\\
    \bottomrule
    \end{tabular}}
    \label{tab:zero_closedie_page}
\end{table*}

\end{document}